\documentclass[runningheads,a4paper]{llncs}

\usepackage{amssymb}
\usepackage{amsmath}
\usepackage{graphicx}
\usepackage{subfig}
\usepackage{multirow}
\usepackage{wrapfig}
\usepackage{floatflt}
\usepackage{arydshln}
\usepackage{graphics}
\usepackage[lined,ruled,linesnumbered]{algorithm2e}

\usepackage{url}
\urldef{\mailsa}\path|{sunil.aryal,a.agraharibaniya}@deakin.edu.au|
\urldef{\mailsb}\path|santosh.kc@usd.edu|

\begin{document}

\mainmatter  


\title{Improved histogram-based anomaly detector with the extended principal component features}

\titlerunning{Improved histogram-based anomaly detector}
%
%

\author{Sunil Aryal$^{1}$, Arbind Agrahari Baniya$^{1}$ and KC Santosh$^{2}$}
\authorrunning{Aryal et. al.}


\institute{$^1$Deakin University, Geelong, Victoria, Australia \\ \mailsa \\ $^2$University of South Dakota, Vermillion, SD, USA \\ \mailsb}
%
%

\maketitle


\begin{abstract}

In this era of big data, databases are growing rapidly in terms of the number of records. Fast automatic detection of anomalous records in these massive databases is a challenging task. Traditional distance based anomaly detectors are not applicable in these massive datasets. Recently, a simple but extremely fast anomaly detector using one-dimensional histograms has been introduced. The anomaly score of a data instance is computed as the product of the probability mass of histograms in each dimensions where it falls into. It is shown to produce competitive results compared to many state-of-the-art methods in many datasets. Because it assumes data features are independent of each other, it results in poor detection accuracy when there is correlation between features. To address this issue, we propose to increase the feature size by adding more features based on principal components. Our results show that using the original input features together with principal components improves the detection accuracy of histogram-based anomaly detector significantly without compromising much in terms of run-time.     


\end{abstract}


\section{Introduction}
\label{sec_intro}

 While majority of records in databases are normal or expected, some of them can be unusual or unexpected. Those data records or instances which are deviated significantly and do not conform to normal data are called \textit{anomalies} or \textit{outliers}. Automatic detection of such anomalies using computers is the task of \textit{Anomaly Detection} (AD) in data mining \cite{TanBook_2006,surveyAD_Chandola2009}. It has many applications such as intrusion detection in computer networks, fraudulent transaction detection in banking and malignant tumor detection in healthcare.
 
 In the data mining literature, the problem of anomaly detection is solved using three learning approaches \cite{surveyAD_Chandola2009}:
 \begin{enumerate}
     \item Supervised: A classification model is learned using training instances from both normal and anomalous classes and the learned model is used to predict the class labels of test data. It requires sufficient labelled training samples from both classes. In many real-world applications, it might be very expensive or even impossible to obtain labelled training samples from the anomalous class.
     \item Unsupervised: Instances in a database are ranked directly based on some outlier score. This approach does not require a training process and labelled samples. Assuming that anomalies are few and different, it uses distance/density to rank them. While this approach works quite well in scenarios where the assumption holds, it results in poor performance when this assumption does not hold, i.e., when  there are far too many anomalies.
     \item Semi-supervised: A profile of normal data is learned using labelled training data from the normal class only. In the testing phase, instances are ranked based on how well they comply with the learned profile of normal data. Neither it makes any assumption about anomalies nor it requires training samples from anomalous class.
     \end{enumerate}
 
Many traditional unsupervised and semi-supervised approaches are based on $k$-Nearest Neighbours ($k$NN) \cite{surveyAD_Chandola2009,LOF_Breunig2000,ORCA_Bay2003} which require pairwise distance calculations. Being few and different anomalies are expected to have large distances to their $k$NNs compared to their neighbours. Because of the computational complexity to compute pairwise distances, these distance based methods are limited to small databases. They can not be used in databases with large number of instances.

In this era of big data, databases are growing rapidly in terms of both the number of instances and dimensionality. Fast automatic detection of  anomalies in these massive databases is a challenging task. There are some efficient unsupervised anomaly detectors \cite{IForest_Liu2008,Sp_Sugiyama2013,hbos_Goldstein2012,SPAD_Aryal2016} which are applicable to large databases. 

Univariate histograms based method \cite{surveyAD_Chandola2009,hbos_Goldstein2012,SPAD_Aryal2016} is arguably the fastest anomaly detector. It creates univariate histograms in each dimension and computes anomaly score of an instance as the product of probability data mass in histograms where the instance falls in all dimensions. It assumes that attributes are independent of each other. Despite the strong assumption, it has been shown to produce competitive results compared to state-of-the-art contenders and runs very fast\cite{hbos_Goldstein2012,SPAD_Aryal2016}. It can detect anomalies that show outlying characteristics in any input features (i.e., one-dimensional subspaces). We refer to such anomalies as `\textit{Type I Anomalies}'. However, it can not detect anomalies which look normal in each one-dimensional subspaces but show outlying characteristics in multidimensional subspaces (we refer to them as `\textit{Type II Anomalies}') as it does not capture the relationship between input features. It may be the case in many applications that there is a correlation between input features and anomalies are not different from normal instances in any single feature but they look different only if features are examined together.

In this paper, we attempt to overcome the above mentioned limitation of histogram based method using principal components \cite{pca_karl1901,PCABook_Jolliffe2005}. Principal Components (PCs) are mappings of subspaces of possibly correlated features in the input space into a new space. We introduce a new variant of histogram based anomaly detection method called `{\tt SPAD+}' where anomaly scores of instances are computed based on univariate histograms of the input features and principal components. This way, histograms in input features contribute to detect Type I anomalies and histograms in PCs contribute to detect Type II anomalies. Our empirical results in 15 benchmark datasets show that SPAD+ significantly improves the detection accuracy of SPAD, a histogram based anomaly detector using input features only, without compromising much in terms of runtimes. It produces better results in terms of both detection accuracy and runtime than traditional nearest neighbour based method and more recent faster anomaly detectors.

Rest of the paper is structured as follows. Preliminaries related to this paper and a review of widely used state-of-the-art fast anomaly detection methods are provided in Section~\ref{sec_relatedWork}. The proposed method of SPAD+ is discussed in Section~\ref{sec_proposed_method} followed by experimental results in Section~\ref{sec_exp} and conclusions in the last section.


\section{Preliminaries and related work}
\label{sec_relatedWork}

In this paper, we focus on continuous domain where each data instance ${\bf x}$ is represented by an $M$-dimensional vector $\langle x_{1},x_{2},\cdots,x_{M}\rangle \in {\mathbb R}^M$ (where ${\mathbb R}$ is a real domain). Each $x_i\in {\mathbb R}$ is the value of the $i^{th}$ feature of ${\bf x}$.
We use semi-supervised approach for anomaly detection where a profile of normal data is learned from training data and test data (a mixture of normal and anomalous data) are ranked according to their anomaly scores based on the learned profile of normal data. Let $D$ be a collection of $N$ training data (all normal) and $Q$ be a collection of $n$ test data (normal and anomalies). The task is to model the profile of normal data from $D$ and ranked data in $Q$ based on their compliance to the profile of normal data.

In the rest of this section we review most widely used and fast anomaly detection methods.

\subsection{Local Outlier Factor (LOF)}

Local Outlier Factor (LOF) \cite{LOF_Breunig2000} is the most widely used and popular $k$NN based anomaly detection method. It does not require any training. The anomaly score of ${\bf x}\in Q$ is calculated based on the distances to its $k$NNs in $D$. It uses the concept of local reachability distance (lrd). Let $kNN({\bf x}, D)$ be the set of $k$NNs of ${\bf x}$ in $D$, $d({\bf x},{\bf y}) = \sqrt{\sum_{i=1}^M (x_i-y_i)^2}$ is the euclidean distance between ${\bf x}$ and ${\bf y}$, and $d^k({\bf x}, D)$ is the euclidean distance between ${\bf x}$ and its $k^{th}$ NN in $D$. The anomaly score of ${\bf x}$ is calculated as:

\begin{equation}
\label{eqn_lof}
s_{lof}({\bf x}) = \frac{\displaystyle \sum_{{\bf y} \in N^k({\bf x}, D)}lrd({\bf y})}{|N^k({\bf x}, D)|\times lrd({\bf x})}
\end{equation}
\noindent $lrd({\bf x}) = \frac{|N^k({\bf x}, D)|}{\sum_{{\bf y} \in N^k({\bf x}, D)} \max\left(d^k({\bf y}, D), d({\bf x}, {\bf y})\right)}$. Note that $|N^k({\bf x}, D)|$ may not be exactly $k$ when there are many instances are equidistant to ${\bf x}$, i.e., $N^k({\bf x}, D)=\{{\bf y}\in D : d({\bf x},{\bf y})\leq d^k({\bf x}, D)\}$.

The anomaly score of $s_{lof}({\bf x})$ is computed with respect to its local neighbourhood in $D$ defined by  $N^k({\bf x}, D)$. It measures how different is ${\bf x}$ with respect to it $k$NNs in terms of lrd. The anomaly score is based on the distances ${\bf x}$ to its $k$NNs and their distances to their $k$NNs. It can be computationally very expensive when $D$ is large limiting its use in small datasets only.

\subsection{Isolation forest (iforest)}

Isolation forest (iforest) \cite{IForest_Liu2008} is an efficient anomaly detector method which does not require pairwise distance calculations. It uses a collection of $t$ unsupervised random trees where each tree $T_i$ is constructed from a small subsamples of training data ${\cal D}_i\subset D, |{\cal D}_i|=\psi\ll N$. The idea is to isolate every sample in ${\cal D}_i$. At each internal node of $T_i$, the space is partitioned into two non-empty regions with a random split on a randomly selected attribute. The partitioning continued until the node has only one instance (which is isolated from the rest) or the maximum height of $\lfloor\log_2\psi\rfloor$ is reached. The anomaly score of ${\bf x}\in Q$ is estimated as the average path length over $t$ trees:

\begin{equation}
\label{eqn_iforest}
s_{if}({\bf x}) = \frac{1}{t}\sum_{i=1}^t \ell_i({\bf x})
\end{equation}
\noindent where $\ell_i({\bf x})$ is the path length of ${\bf x}$ in tree $T_i$.

The simple intuition is that anomalies are more susceptible to isolation and they have shorter average path lengths than normal instances. It is shown to produce competitive detection accuracy to LOF but runs significantly faster as it does not require distance calculations \cite{IForest_Liu2008}.

\subsection{Nearest neighbour distance in a small subsample (Sp)}

Instead of using $k$NNs of ${\bf x}$ in the entire training data $D$, Sugiyama and Borgwardt (2013) \cite{Sp_Sugiyama2013} argued that it is sufficient to use 1NN ($k=1$) in a small subsample of data ${\cal D}\subset D, |{\cal D}|=\psi\ll N$. They suggested to use the distance of ${\bf x}$ to it nearest neighbour in ${\cal D}$ as the anomaly score.

\begin{equation}
\label{eqn_sp}
s_{sp}({\bf x}) = \displaystyle \min_{{\bf y}\in {\cal D}} d({\bf x},{\bf y})
\end{equation}

It has been shown that $\psi$ as small as $20$ or $25$ produces competitive results to LOF but runs orders of magnitude faster in large datasets \cite{Sp_Sugiyama2013}.

\subsection{Histogram based method}

Another simple and efficient anomaly detection method is based on univariate histograms \cite{hbos_Goldstein2012}. In each dimension $i$, it creates histograms with a fixed number of equal-width bins $h_{i,1},h_{i,2},\cdots,h_{i,b}$ and records the number of training instances falling in each $h_{i,j}$. Anomalies are expected to fall in bins with small number of training samples. 

Aryal et al. (2016) introduced a simple probabilistic anomaly detector (SPAD) \cite{SPAD_Aryal2016} where multivariate probability of ${\bf x}$, $P({\bf x})$, is approximated as the product of univariate probabilities $P(x_i)$ assuming attributes are independent of each other. Approximation of $P(x_i)$ is estimated using probability mass by discretising values in dimension $i$ based on equal-width histograms. They use a modified version of equal-width discretisation which is more robust to skewed distribution of data in dimension $i$. Instead of dividing the entire data range defined by $[\min_i, \max_i]$ ($\min_i$ and $\max_i$ are the minimum and maximum value in dimension $i$) into $b$ equal-width bins, they divide the range $[\mu_i-3\sigma_i, \mu_i+3\sigma_i]$ ($\mu_i$ and $\sigma_i$ are the mean and standard deviation of values in dimension $i$) into $b$ equal-width bins. The bin width in each dimension depends on the data variance in that dimension. The anomaly score of ${\bf x}$ is then estimated as:

\begin{equation}
\label{eqn_scoreSPAD}
s_{spad}({\bf x}) = \displaystyle \sum_{i=1}^M \log{\frac{|H_{i}({\bf x})|+1}{N+b}}
\end{equation}

\noindent $H_{i}({\bf x})$ is the bin in dimension $i$ where ${\bf x}$ falls into. Note that the RHS of Eqn~\ref{eqn_scoreSPAD} is equivalent to the logarithm of $\hat{P}({\bf x})=\prod_{i=1}^M \hat{P}(x_i)$.

Simple univariate histogram based method is shown to produce competitive results to LOF, iforest and Sp and it runs faster than them \cite{hbos_Goldstein2012,SPAD_Aryal2016,usfAD_Aryal2018}.

Among the four anomaly detection methods reviewed above, LOF is a widely used baseline and other three methods are fast anomaly detectors for large datasets. Their time and space complexities are provided in Table~\ref{tbl_complexity}. In terms of time and space complexities, SPAD is clearly the most efficient one.

\begin{table}[t]
\centering
\caption{Time (training and testing) and space complexities. Note that $N$:\#training data, $M$:\#dimensions, $n$:\#test data, $\psi$:\#subsamples, $t$:\#trees and $b$:\#bins}
\begin{tabular}{ l | r @{\hspace{20pt}} r @{\hspace{5pt}} | r }
\hline
\multirow{2}{*}{Method\mbox{  }} & \multicolumn{2}{c|}{Time} & \multirow{2}{*}{Space} \\
\cline{2-3}
   & Train & Test & \\
\hline
LOF & \mbox{   } $O(N\log_2NM)^\dagger$ & $O(n\log_2NM)$ & \mbox{   }$O(NM)$ \\
iforest & \mbox{   }$O(t\psi\log_2\psi)$ & $O(nt\log_2\psi)$ & $O(t\psi)$ \\
Sp & - & $O(nM\psi)$ & $O(M\psi)$ \\
SPAD & $O(NMb)$ & $O(nM)$ & $O(Mb)$ \\
\hline
\end{tabular}
\begin{flushleft}
\vspace{-5pt}
\hspace{1.6cm}$^\dagger$Not training per se, it is to compute $lrd(\cdot)$ of all training instances\\
\hspace{1.6cm}to use later while computing $s_{lof}(\cdot)$ for test instances.
\end{flushleft}
\label{tbl_complexity}
\end{table}

\section{The proposed method: SPAD+}
\label{sec_proposed_method}

\begin{figure}[t]
\centering
\subfloat[Input space]{\includegraphics[width=0.48\textwidth]{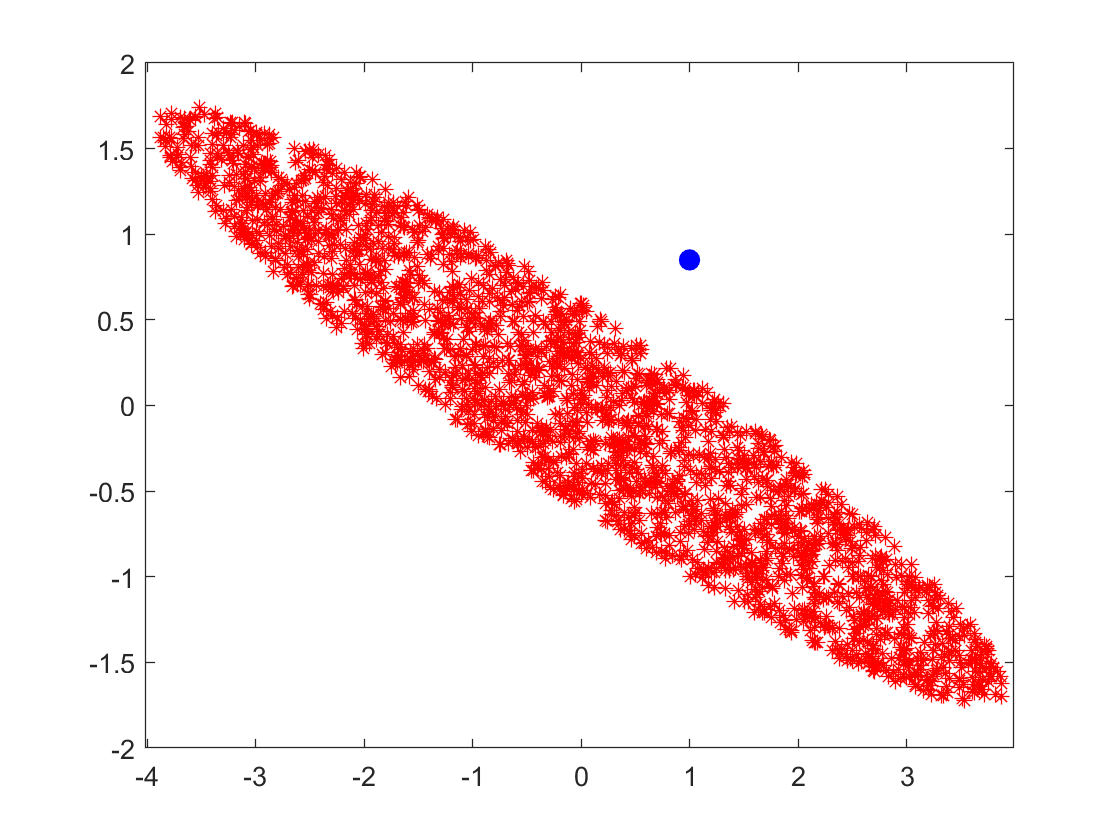}}
\subfloat[PC space]{\includegraphics[width=0.48\textwidth]{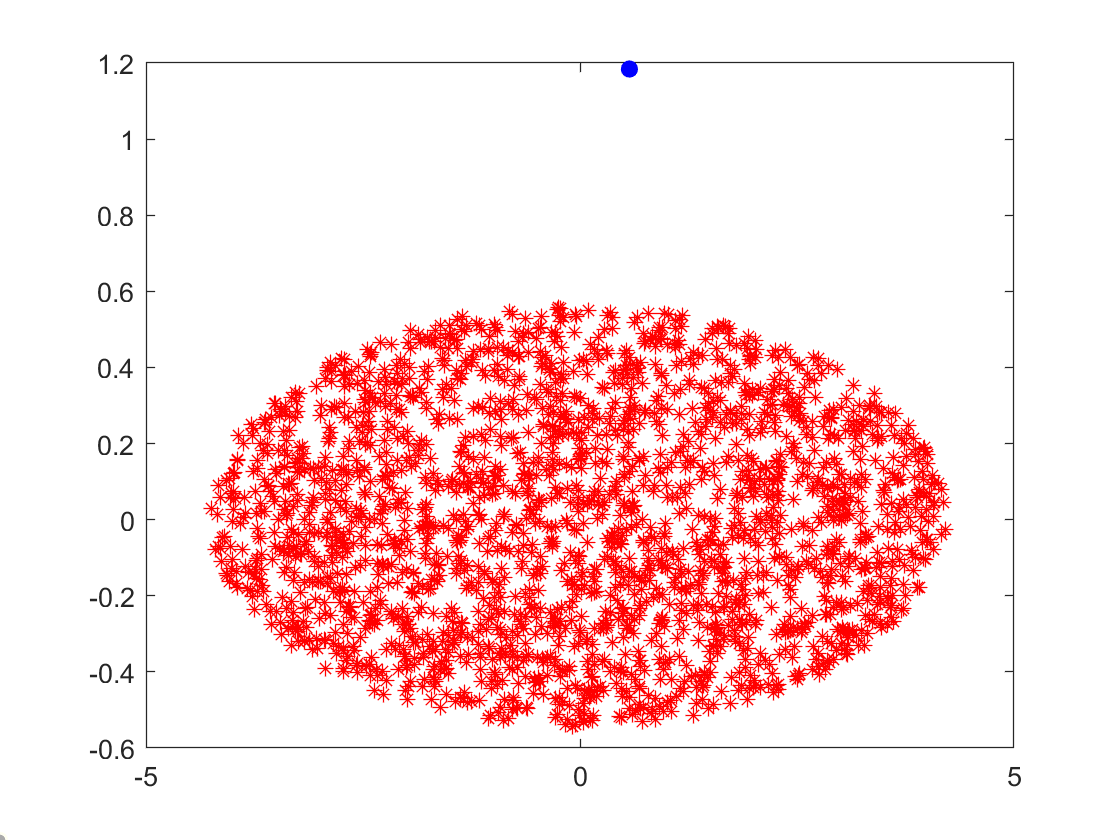}}
\caption{An example scenario (a) SPAD fails to detect anomaly (on blue) in the input space and (b) SPAD can easily detect the anomaly in the space defined by the principal components.}
\label{fig_example_data}
\end{figure}

Though SPAD can detect anomalous data exhibiting outlying characteristics in any dimension (Type I Anomalies), it can not detect anomalous data which look normal in each dimension but exhibit outlying characteristics only when examined on multiple features together (Type II Anomalies) \cite{SPAD_Aryal2016}. For example, the data point shown on blue in Fig~\ref{fig_example_data}(a) looks perfectly normal when examined from each dimension individually (it is in the middle of the distribution in each dimension). SPAD can not detect such obvious anomaly. This is because it does not capture the relationship between input features. In many real-world applications, features can be related and anomalous data may not conform to it.

We believe the above mentioned issue of SPAD can be addressed to some extent with Principal Component Analysis (PCA)\cite{pca_karl1901,PCABook_Jolliffe2005}. PCA is a tool to transform potentially correlated features into new uncorrelated features \cite{PCABook_Jolliffe2005}. It learns the transformation matrix using the covariance matrix of observed data. Fig~\ref{fig_example_data}(b) is the transformation of data in Fig~\ref{fig_example_data}(a) in the PC space. The anomalous instance on blue is clearly an outlier in the second principal component (vertical axis in the Fig~\ref{fig_example_data}(b)) which can be easily detected by SPAD in the new PC space.

In order to cater for both Type I and Type II of anomalies, we propose to add principal components (PCs) as new features in addition to the input features. Input features  contribute  to  detect Type  I  anomalies and PCs contribute to detect Type II anomalies. Using PCs only may not be a good idea because it may mask anomalies which are easily detectable in the input space\footnote{In our experiments, we observed that using PCs only produced worst results in many datasets.}. 

For semi-supervised anomaly detection, PCA is applied to the training dataset $D$ to learn the transformation matrix which is used to transform data in both $D$ and $Q$. The feature size is increased from $M$ to $2M$ by adding all $M$ PCs. In SPAD, histograms are constructed in each dimension of the input space and PC space. We refer the SPAD used in input features and PCs as {\tt SPAD+}. Let ${\bf x}'$ be the transformation of ${\bf x}$ in the PC space. The anomaly score of ${\bf x}\in Q$ is estimated as:

\begin{equation}
\label{eqn_scoreSPAD+}
s_{spad+}({\bf x}) = \displaystyle \sum_{i=1}^M \log{\frac{|H_{i}({\bf x})|+1}{N+b}} + \displaystyle \sum_{j=1}^M \log{\frac{|H_{j}({\bf x}')|+1}{N+b}}
\end{equation}

In the literature, PCA is used mainly to reduce dimensionality of data where top $m<M$ PCs that capture the most variance in data are selected and PCs with low variance are ignored. However, the purpose of using PCs in SPAD+ is not to reduce dimensionality, they are used to capture correlation between attributes to detect Type II anomalies. Low variance PCs can be very useful to detect anomalies as they may contain few values which are significantly different from the rest, more likely to be anomalies (e.g., in Fig!\ref{fig_example_data}(b), the blue point is an outlier in the second PC where variance is lower than in the first PC). Ignoring low variance PCs can be counter productive for anomaly detection\footnote{In our experiments, we observed that adding top PCs capturing 95\% variance of data produced worse results than adding all PCs in many datasets.}. Thus, we add all PCs so that maximum possible anomalies are detected.   

In terms of runtime and space complexities, SPAD+ has similar complexities as SPAD. It requires additional $O(NM+M^3)$ time (worst case) in the training process to compute the covariance matrix of the training data $D$ and its eigen decomposition. In the testing phase the only difference is that $M$ is increased to $2M$. Because of $O(M^3)$, it can be computationally expensive in high dimensional datasets where $M$ is large. However, it is the worst case and on average case it can be done faster. It may not be an issue unless $M$ is very large in thousands or millions.


\section{Empirical evaluation}
\label{sec_exp}

In this section, we provide details of our experimental setup to evaluate performances of SPAD+ against the four methods (LOF, iforest, Sp and SPAD) discussed in Section~\ref{sec_relatedWork} and discuss results. We conducted experiments in semi-supervised setting --- half of the data belonging to the normal class were used as training set $D$ and the remaining half along with data belonging to the anomaly class were used as test set $Q$ as done in \cite{catSimLOF_Boriah2008}. Anomaly detection model was learned from the training data and test data were ranked based on their anomaly score using the learned model. We used Area under the receiver operation curve (AUC) as the performance evaluation measure. For the random methods - iforest and Sp, each experiment was repeated 10 times and reported the average AUC over 10 runs. For a dataset, the same $D$ and $Q$ pair were used for all experiments. Min-max normalisation\cite{TanBook_2006} was used to ensure feature values in each dimension are in the same range. 

All methods were implemented in Python using the Scikit-learn machine learning library \cite{scikit-learn_2011}. We used the LOF implementation available on the Scikit-learn library. All experiments were conducted on a Linux machine. Parameters in all algorithms were set to default values suggested by respective papers --- LOF ($k=\lfloor\sqrt N\rfloor$); iforest ($t=100$ and $\psi=256$); Sp ($\psi=25$); and $b=\lfloor\log_2N\rfloor+1$ in SPAD and SPAD+.

We used 15 widely used publicly available benchmark datasets from various application areas such as space physics, health and medicine, cyber security, pharmaceutical chemistry and geographical information system (GIS). The characteristics of datasets in terms of dimensionality, training data size, the numbers of normal data and anomalies in the test data and application area are provided in Table~\ref{tbl_data}.  

\begin{table}[t]
\centering
\caption{Benchmark datasets}
\begin{tabular}{ l @{\hspace{10pt}} r @{\hspace{10pt}} r @{\hspace{10pt}} r @{\hspace{10pt}} r @{\hspace{20pt}} l }
\hline
 \multirow{2}{*}{Name} & \#dim & Training & \multicolumn{2}{c}{Test data} & \mbox{     }Application\\
\cline{4-5}
                     & ($M$) & size ($N$) & \#Normal & \#Anomalies & \mbox{     }Area \\  
\hline
Ionosphere\cite{UCI_Dua2017} &  32 &  112 &  113 &  126 & \mbox{     }Space \\
Dermatology\cite{UCI_Dua2017} &  33 &  173 &  173 &  20 & \mbox{     }Health \\
Arrhythmia\cite{UCI_Dua2017} &  274 &  193 &  193 &  66 & \mbox{     }Health \\
Pima\cite{UCI_Dua2017} &  8 &  250 &  250 &  268 & \mbox{     }Health \\
Spambase\cite{UCI_Dua2017} &  57 &  1394 &  1394 &  176 & \mbox{     }Cyber \\
Musk2\cite{UCI_Dua2017} &  166 &  2790 &  2791 &  291 & \mbox{     }Chemistry \\
Satellite\cite{UCI_Dua2017} &  36 &  2199 &  2200 &  2036 & \mbox{     }Image \\
Annthyroid\cite{UCI_Dua2017} &  21 &  3333 &  3333 &  534 & \mbox{     }Health\\
Phishing\cite{kagle_datasets} &  30 &  3078 &  3079 &  4898 & \mbox{     }Cyber \\
ISCXURL\cite{unbcic_datasets} &  75 &  3889 &  3890 &  7570 & \mbox{     }Cyber \\
Mnist\cite{kagle_datasets} &  96 &  9884 &  9884 &  676 & \mbox{     }Image\\
U2R\cite{UCI_Dua2017} &  33 &  30296 &  30297 &  228 & \mbox{     }Cyber \\
NSL-KDD\cite{unbcic_datasets} &  38 &  38527 &  38527 &  71463 & \mbox{     }Cyber \\
UNSW\cite{unswnb15_Moustafa2015} &  39 &  82336 &  82337 &  93000 & \mbox{     }Cyber \\
Covertype\cite{UCI_Dua2017} &  10 &  141650 &  141651 &  2747 & \mbox{     }GIS\\
\hline
\end{tabular}
\label{tbl_data}
\end{table}

\begin{table}[!htb]
\centering
\caption{Anomaly detection performance (AUC). The best result in each dataset is bold faced. The average AUC and average rank of each method over the 15 datasets are provided in the last two rows.}
\begin{tabular}{@{\hspace{5pt}} l @{\hspace{20pt}} l @{\hspace{20pt}} l @{\hspace{22pt}} l @{\hspace{22pt}} l @{\hspace{20pt}} l @{\hspace{5pt}}}
\hline
  Name & LOF & iforest & Sp & SPAD & SPAD+ \\
\hline 
Ionosphere & \textbf{0.9671} &  0.9032 &  0.9624 &  0.7208 & 0.9475 \\
Dermatology &  \textbf{1.0000}&  0.9567 &  0.9997 &  0.8038 & 0.9908 \\
Arrhythmia &  0.8267 &  0.8301 &  0.8312 &  0.8231 & \textbf{0.8372} \\
Pima &  0.7067 &  0.7492 &  0.7276 &  0.7427 & \textbf{0.7626}\\
Spambase &  0.7240 &  0.8091 &  0.7379 &  0.7703 & \textbf{0.8318}\\
Musk2 &  0.6993 &  0.4267 &  0.5526 &  0.6044 & \textbf{0.7834}\\
Satellite &  0.8370 &  0.7929 &  0.8437 &  \textbf{0.8676}& 0.8648 \\
Annthyroid &  0.7493 &  0.7416 &  0.6359 &  \textbf{0.8701}& 0.7884 \\
Phishing &  \textbf{0.7200}&  0.6367 &  0.5795 &  0.6375 & 0.6584 \\
ISCXURL &  \textbf{0.9040}&  0.8419 &  0.8366 &  0.8027 & 0.8768 \\
Mnist &  \textbf{0.8789}&  0.8352 &  0.8140 &  0.7988 & 0.8587 \\
U2R &  0.8869 &  0.9865 &  0.9834 &  0.9771 & \textbf{0.9929}\\
NSL-KDD &  0.9000 &  0.9613 &  0.9486 &  0.9355 & \textbf{0.9625}\\
UNSW &  0.8506 &  0.8347 &  0.8849 &  0.8181 & \textbf{0.9042}\\
Covertype &  \textbf{0.9929}&  0.8480 &  0.8900 &  0.8279 & 0.9328 \\
\hline
Avg. AUC &  0.8429 &  0.8102 &  0.8152 &  0.8000 & \textbf{0.8662} \\
Avg. rank & 2.80 & 3.40 &  3.33  &  3.80  &  1.67 \\
\hline
\end{tabular}
\label{tbl_auc}
\end{table}

The AUC of all five contenders in the 15 datasets are provided in Table~\ref{tbl_auc}. SPAD+ produced the best results in seven datasets, while producing the second best results in six datasets and the third best results in the remaining two datasets. It was ranked among the top three best performing methods in all the 15 datasets. The baseline method of LOF is the closest contender with the best AUC in six datasets. It was ranked second in one dataset, third in three datasets and fourth in four datasets. It was the worst performing method in the remaining four datasets. Sp and iforest did not produce the best result in any dataset and each of them produced the worst results in two datasets. While original SPAD produced the best result in two datasets, it produced the worst results in seven datasets. 

From this results, it is clear that adding PCs, SPAD+ significantly improved the AUC in 13 out of 15 datasets, making it the top performing method from the worst. This is because of the ability to examine anomalies in individual input features and multidimensional subspaces represented by PCs which enable SPAD+ to detect both Type I and Type II anomalies. Existing fast anomaly detectors' results are not comparable to those of SPAD+. iforest did not produce better AUC than SPAD+ in any dataset whereas Sp produced better AUC than SPAD+ in two datasets only. Compared to SPAD, SPAD+'s was slightly worse in Satellite and was significantly worse in Annthyroid but it was still better than all other contenders. 

\begin{table}[t]
\centering
\caption{Total runtime including training and testing in seconds.}
\begin{tabular}{@{\hspace{5pt}} l @{\hspace{20pt}} r @{\hspace{20pt}} r @{\hspace{22pt}} r @{\hspace{22pt}} r @{\hspace{20pt}} r @{\hspace{5pt}}}
\hline
  Name & LOF & iforest & Sp & SPAD & SPAD+ \\
\hline 
Ionosphere &  0.25 &  2.48 &  0.16 &  0.32 & 3.35 \\
Dermatology &  0.25 &  2.22 &  0.15 &  0.28 &  0.71 \\
Arrhythmia &  0.46 &  2.16 &  0.28 &  1.74 &  2.27 \\
Pima &  0.49 &  2.91 &  0.28 &  0.15 &  0.31 \\
Spambase &  2.12 &  3.70 &  1.00 &  2.39 & 11.83 \\
Musk2 &  12.87 &  8.83 &  2.14 &  12.54 & 19.25 \\
Satellite &  6.02 &  12.07 &  2.73 &  2.00 &  7.42 \\
Annthyroid &  3.61 &  5.54 &  1.95 &  1.73 & 2.75 \\
Phishing &  21.76 &  11.96 &  4.11 &  4.90 &  4.99 \\
ISCXURL &  72.20 &  21.74 &  40.16 &  12.77 & 18.17 \\
Mnist &  160.36 &  27.94 &  8.21 &  34.70 & 67.19 \\
U2R &  229.78 &  55.54 &  19.09 &  43.24 &  74.62 \\
NSL-KDD  &  66.58 &  514.55 &  35.35 &  34.56 & 67.20 \\
UNSW &  398.69 &  613.74 &  316.23 &  114.33 & 184.65 \\
Covertype &  438.21 &  150.15 &  50.85 &  26.49 & 44.13 \\
\hline
\end{tabular}
\label{tbl_runtime}
\end{table}

The total runtimes (training and testing) of the five contending measures in the 15 datasets are provided in Table~\ref{tbl_runtime}. As expected SPAD+ ran slower than SPAD. Compared to other contenders, though it was slower in small datasets, it ran faster than them in large datasets. For example, SPAD+ was one order of magnitude faster than LOF in the largest dataset.   

These results show that SPAD+ significantly improves anomaly detection performance without compromising much in terms of runtime, particularly in large datasets with a large number of data instances. It is a simple and intuitive method which is more appropriate for big data characterised by large data size. 

The only difference between SPAD and SPAD+ is the addition of PCs as new features. One can argue that we can use the same idea with existing methods. We tested using the original input features and PCs and the results are presented in Table~\ref{tbl_auc_pca}. The results of LOF and Sp remained largely unchanged whereas that of iforest were improved in some datasets. iforest managed to produce better AUC than SPAD+ in Ionosphere only, it's results in other 14 datasets were worse than those of SPAD+.

\begin{table}[t]
\centering
\caption{Anomaly detection performance (AUC) in benchmark datasets in the space made up of input features + Principal Components (PCs). The average AUC over the 15 datasets is provided in the last row.}
\begin{tabular}{@{\hspace{5pt}} l @{\hspace{20pt}} l @{\hspace{20pt}} l @{\hspace{22pt}} l @{\hspace{22pt}} l @{\hspace{5pt}}}
\hline
  Name & LOF & iforest & Sp & SPAD+ \\
\hline 
Ionosphere & \textbf{0.9671}&  0.9559 &  0.9624 & 0.9475 \\
Dermatology &  \textbf{1.0000}&  0.9832 &  0.9997 & 0.9908 \\
Arrhythmia &  0.8271 &  0.8279 &  0.8316 & 0.\textbf{8372}\\
Pima & 0.7067 &  0.7555 &  0.7276 & \textbf{0.7626}\\
Spambase & 0.7240 &  0.8125 &  0.7379 & \textbf{0.8318}\\
Musk2 & 0.6993 &  0.6962 &  0.5526 & \textbf{0.7834}\\
Satellite & 0.8370 &  0.7814 &  0.8437 & \textbf{0.8648}\\
Annthyroid & 0.7493 &  0.6925 &  0.6359 & \textbf{0.7884}\\
Phishing & \textbf{0.7236}&  0.6306 &  0.5883 &  0.6584 \\
ISCXURL &  \textbf{0.9040}&  0.8657 &  0.8366 & 0.8768 \\
Mnist &  \textbf{0.8789}&  0.8500 &  0.8140 & 0.8587 \\
U2R & 0.8875 &  0.9904 &  0.9834 &  \textbf{0.9929}\\
NSL-KDD & 0.9000 &  0.9578 &  0.9478 & \textbf{0.9625}\\
UNSW & 0.8507 &  0.8875 &  0.8849 & \textbf{0.9042}\\
Covertype & \textbf{0.9929}&  0.9233 &  0.8900 & 0.9328 \\
\hline
Avg. AUC &  0.8432 &  0.8407 &  0.8158 & \textbf{0.8662}\\
\hline
\end{tabular}
\label{tbl_auc_pca}
\end{table}

Adding PCs did not add any value to LOF and Sp. It is because of the anomaly scores used which are based on nearest neighbours distances. They are already examining anomalies using all dimensions even when only input features are used. Adding PCs results in additional redundant feature and does not necessarily impact the outcome of the algorithm.  

However, PCs can be useful in iforest. When only input features are used, it can examine anomalies using multiple dimensions to some extent because leaf nodes in trees are constructed by partitioning space using different attributes. However, the number of attributes used in the examination are restricted by the height of the leaf nodes. In best case, it can use up to 8 ($\log_2 256$ as $\psi$ was to 256 by default) dimensions. Because attribute is selected at random and tree building process terminates early when a sample instance is isolated, even 8 attributes may not be used. Therefore adding PCs enables it to examine anomalies using many attributes at the same time to detect more Type II anomalies. Because iforest mostly examines anomalies using randomly selected multiple attributes, it forces correlation even though it may not exist in the dataset which results in false positive --- some normal instances may appear to be anomalies. This could possibly be one of the reasons why iforest can not perform as good as SPAD+ when input features and PCs are used because PCs by definition are uncorrelated, i.e., they are orthogonal to each other. 

In SPAD, PCs are very useful to examine anomalies using potentially correlated attributes together to detect Type II anomalies. Because it examines each dimension individually and does not force any non existing correlation, it can avoid false positive cases like those in iforest. In SPAD+, original input features can be useful to detect Type I anomalies and PCs can be useful to detect Type II anomalies, complimenting each other to produce better outcome when applied together.



\section{Concluding remarks}
\label{sec_con}

The idea of estimating multivariate probability as the product of univariate probabilities assuming that variables are independent of each other is widely used in other data mining and machine learning tasks such as naive Bayes classifier \cite{TanBook_2006}. It has not been explored enough in the anomaly detection task. Histogram based method such SPAD are simple, intuitive and very fast for anomaly detection. It should be a simple baseline to compare the performances of more complex and recent approaches. However, only a few studies such as \cite{hbos_Goldstein2012,SPAD_Aryal2016} are using it.

Even though SPAD produces results quite competitive to other efficient anomaly detectors, it has a limitation on detecting anomalies which rely on multiple attributes because it examines each attribute separately assuming that they are independent from each other \cite{SPAD_Aryal2016}. In this paper, we show that this limitation can be addressed to some extent by using principal components (PCs) of data as additional features. The idea is to double the feature size with input features and PCs.Then, use SPAD in the new space. We call the new variant of SPAD using PCs as {\tt SPAD+}. Our empirical results show that SPAD+ significantly improves the performance of SPAD without compromising much in runtime and results in better performance than state-of-the-art methods. It runs faster than other existing fast anomaly detection methods. 

It's simplicity, effectiveness and efficiency make SPAD+ an ideal anomaly detector to use in big data with hundreds of thousands to millions of data instances. We believe this simple method sets a new baseline for anomaly detection performance comparison. 

PCs based on the covariance matrix only capture linear relationships between input features. They can not capture non-linear relationships between features. Kernel-PCA \cite{PCABook_Jolliffe2005} that uses kernel matrix can capture the non-linear relationships but it requires calculations of pairwise kernel similarities of data instances. It is computationally very expensive, limiting its use in small datasets only. We are looking forward to investigating how non-linear relationships present among different attributes in a dataset can be captured efficiently; thereby, improvising the current performance of SPAD+ algorithm. 




\bibliographystyle{splncs}
\bibliography{spad+}

\end{document}